\documentclass[letterpaper]{article}
\usepackage[margin=1in]{geometry}
\usepackage{graphicx,amsmath,amsthm,amssymb}
\usepackage{natbib}
\usepackage{url,color,booktabs}
\usepackage{hyperref}

\newcommand{\C}{\ensuremath{\mathbf{C}}}

\newcommand{\E}{\ensuremath{\mathbf{E}}}

\newcommand{\G}{\ensuremath{\mathbf{G}}}

\newcommand{\X}{\ensuremath{\mathbf{X}}}

\renewcommand{\c}{\ensuremath{\mathbf{c}}}

\newcommand{\sss}{\ensuremath{\mathbf{s}}}  

\newcommand{\w}{\ensuremath{\mathbf{w}}}
\newcommand{\x}{\ensuremath{\mathbf{x}}}

\newcommand{\blambda}{\ensuremath{\boldsymbol{\lambda}}}

\newcommand{\bbR}{\ensuremath{\mathbb{R}}}

\newcommand{\calL}{\ensuremath{\mathcal{L}}}

\newcommand{\calR}{\ensuremath{\mathcal{R}}}

\newcommand{\norm}[1]{\left\lVert#1\right\rVert}

{%
\begin{list}{#1}{
\vspace{-\topsep}
\vspace{-\partopsep}
\setlength{\itemindent}{0cm}
\setlength{\rightmargin}{0cm}
\setlength{\listparindent}{0cm}
\settowidth{\labelwidth}{#1}
\setlength{\leftmargin}{\labelwidth}
\addtolength{\leftmargin}{\labelsep}
\setlength{\itemsep}{0cm}
}%
}%
{%
\end{list}
\vspace{-\topsep}
\vspace{-\partopsep}
}

{\begin{enumerate}%
}%
{\end{enumerate}}

\DeclareMathOperator*{\argmin}{arg\,min}
\DeclareMathOperator*{\argmax}{arg\,max}

\graphicspath{{grf/},{grf_supp/}}

\bibpunct[, ]{(}{)}{;}{a}{,}{,} 

\title{Sampling the ``Inverse Set'' of a Neuron:\\ An Approach to Understanding Neural Nets}
\author{
Suryabhan Singh Hada $\hspace{2ex}$ Miguel {\'A}.\ Carreira-Perpi{\~n}{\'a}n\\
Dept. CSE, University of California, Merced\\
{\url{http://eecs.ucmerced.edu}}
}
\date{December 24, 2020}

\definecolor{DarkRed}{rgb}{0.545,0,0}
\hypersetup{
breaklinks=true, colorlinks=true, linkcolor=black,
citecolor=black, urlcolor=DarkRed 
}

\begin{document}

\maketitle
\begin{abstract}
  With the recent success of deep neural networks in computer vision, it is important to understand the internal working of these networks. What does a given neuron represent? The concepts captured by a neuron may be hard to understand or express in simple terms. The approach we propose in this paper is to characterize the region of input space that excites a given neuron to a certain level; we call this the \emph{inverse set}. This inverse set is a complicated high dimensional object that we explore by an optimization-based sampling approach. Inspection of samples of this set by a human can reveal regularities that help to understand the neuron. This goes beyond approaches which were limited to finding an image which maximally activates the neuron \citep{Simony_14a} or using Markov chain Monte Carlo to sample images \citep{Nguyen_17a}, but this is very slow, generates samples with little diversity and lacks control over the activation value of the generated samples. Our approach also allows us to explore the intersection of inverse sets of several neurons and other variations.	
\end{abstract}

\section{Introduction}
\label{intro}
Recently, deep neural networks have shown great results in solving problems in the field of computer vision. This leads to a surge in the usage of deep neural networks in real life applications, which makes it very important to understand the working of deep neural networks. Some machine learning models are readily interpretable, that is, by looking at the parameters and structure of the model, we can understand the prediction for a given input. For instance, in the case of the univariate decision tree, the prediction for a given input can be written as a sequence of single-feature tests. Another example is the nearest-neighbor classifier, where the model is the training data, and the prediction is given by the most similar training data to the input. However, this is not the case with deep neural networks, where the parameters of the model are interleaved in a very complex way. It is very difficult to make the prediction just by looking at the parameters of the model. In this paper, we propose a general approach to interpret deep neural networks and other complex machine learning models. We achieve this by characterizing the region of input space that excites a given neuron to a certain level; we call this the \emph{inverse set}. That is, rather than looking at the model directly, we propose to treat the model as a black box and characterize its observed behavior on data.

In biology, one puts an electrode into the neuron (Electrophysiology) and records its response while a number of stimuli are fed to it. This is the classic Hubel and Wiesel experiment. We can replicate the same experiment by passing the training data to the network and record its activation for the neuron of interest. However, there are issues with it, as there is no guarantee that we know the data on which network is trained. Even if we know the training data, this approach will never characterize the input space accurately, because the training data is limited. This leads to incomplete information about the nature of neuron. Besides, real-life images contain a lot of irrelevant data, so there is no way to tell which part of the image is important to the neuron. Therefore, we will characterize our input space with synthetic inputs.

\begin{figure*}[t!]
  \centering
  \begin{tabular}{@{}l@{}l@{}}
    \rotatebox{90}{\hspace{.3ex}{act:[50,60]}}&
    \includegraphics*[width=.98\linewidth]{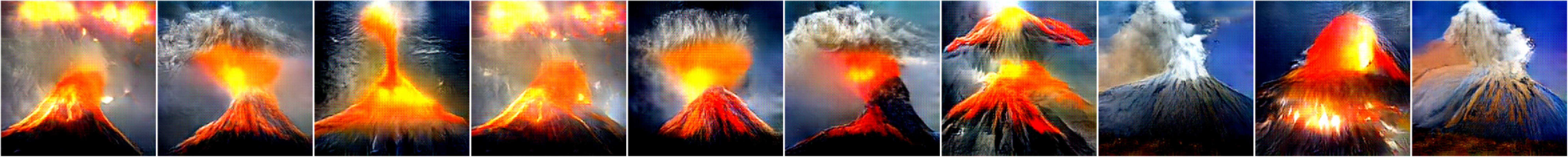}\\\\
    \rotatebox{90}{\hspace{.3ex}{act:[40,50]}}&
    \includegraphics*[width=.98\linewidth]{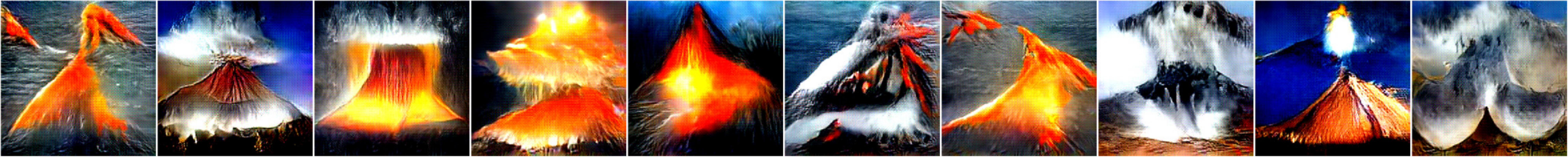}\\\\
    \rotatebox{90}{\hspace{.3ex}{act:[30,40]}}&
    \includegraphics*[width=.98\linewidth]{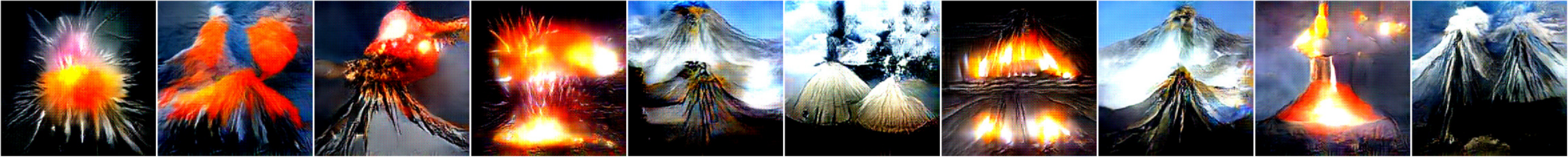}\\\\
    \rotatebox{90}{\hspace{.3ex}{act:[20,30]}}&
    \includegraphics*[width=.98\linewidth]{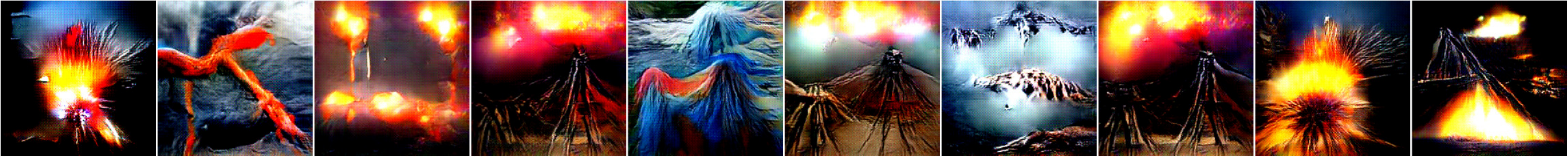}\\\\
    \rotatebox{90}{\hspace{.3ex}{act:[10,20]}}&
    \includegraphics*[width=.98\linewidth]{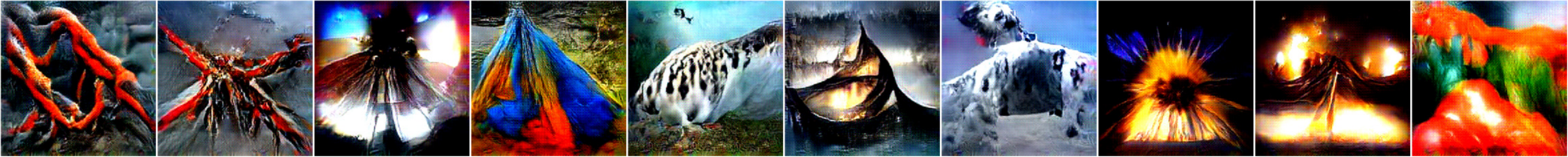}\\\\
    \rotatebox{90}{\hspace{.8ex}{act:[1,11]}}&
    \includegraphics*[width=0.98\linewidth]{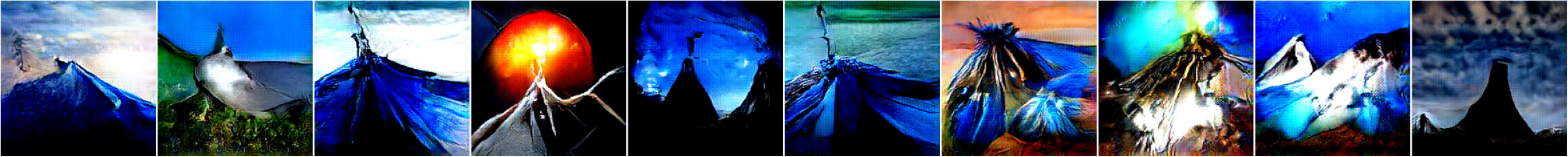}
  \end{tabular}
  \caption{Samples generated using our sampling approach for the neuron number $981$ in the fc8 layer of the CaffeNet \citep{Jia_14a}, which represents volcano class. Each row contains $10$ samples picked from $500$ samples generated with activation range mentioned on the left side. Generated samples are not just photo-realistic but also very diverse in nature, characterizing input space around certain activation very well. Unlike previous approaches~\citep{Nguyen_16b,Nguyen_17a}; generated samples not just only contains volcanoes but they also contain lava flowing through water and on land, and the ash cloud after the volcanic eruption. These kind of samples never been seen before. The first row that contains samples of high activation and the last row which have samples with low activation, both have volcanoes but the presence of lava and smoke create a huge difference in activation value of the neuron. Giving us a clear understanding how the amount of the lava and smoke impact the activation of the neuron, higher the amount of lava and smoke more the activation.}
  \label{f:fc8_1_1_981_sample}
\end{figure*}  

\section{Related work}
\label{related_work}
In a broader sense, we can divide the deep net interpretability research into two categories: 1) produce an explanation in the input space that explains network prediction for a given input, and 2) understand what information is retained by the network.

In the first category, we want to understand the network decision specific to a given instance; it is called \emph{local explanation }~\citep{Guidot_18a}. Methods like \citet{ribeir_16a} and \citet{Guidot_19a} approximate the deep net locally with a much similar model like decision trees, or linear model; and using these approximate models, they provide an explanation in terms of input segments (like superpixels in case of images). Other works \citep{Simony_14a,ZeilerFergus14a,Zhou_16c,Selvar_17a,FongVedald17a} create a saliency maps to explain the network decision. Another line of work finds the training data points that influence the network decision for a given input \citep{KohLiang17a,Yeh_18a}.

Our work belongs to the second category, where we want to understand what information is retained by the network. This provides a global explanation, i.e. the explanations are independent of the given input. Here, we focus only on the image data, but our work can easily be extanded to other data types.

In the last few years, significant effort has been made to understand what parameters of deep neural networks have learned. Currently available techniques that are used to understand the parameters of a trained deep neural network involve two major approaches: feature inversion and activation maximization. Feature inversion attempts to project the output of a layer to the input space, to understand what each layer of the network has learned. This is done by back propagation, gradient descent or training a neural network \citep{MahendVedald15a,Wei_15a,ZeilerFergus14a,DosovitBrox16a}.
\citet{Zhiqiang_16a} used feature inversion to visualize what every filter has learned.

On the other hand, activation maximization involves maximizing the response of a neuron in deep neural network for an input image, as done by \citet{Erhan_09b,Simony_14a,Yosinski_15a,Nguyen_16a, Nguyen_16b}, and many more. Activation maximization is done by taking a random image and then back propagating it through the network to maximize the activation of the neuron of interest.

Both feature inversion and activation maximization suffer from a major problem: the result is a very noisy image or fooling image \citep{Nguyen_15a,Szeged_14a}, which mostly does not make sense to humans. So, to tackle this problem many handcrafted regularizers are used like $\alpha$-norm \citep{Simony_14a}, Gaussian blur \citep{Yosinski_15a}, total variation \citep{MahendVedald15a},
jitter \citep{Mordvintsev_15a}, data-driven patch priors \citep{Wei_15a}, and center-biased regularization \citep{Nguyen_16b}, but there are more. All these methods have helped to improve the quality of the generated images, but still not enough to create natural-looking images. Recently, \citet{DosovitBrox16b} trained a neural network using GAN \citep{Goodfel_14a} to generate realistic images from a feature vector, which is later used by \citet{Nguyen_16a} to generate realistic images for activation maximization.

Although these approaches are really good, they do not consider the fact that there are multiple images which can maximally activate the same neuron. \citet{Nguyen_16b} tried to address this problem by initializing the input image with the mean of a cluster of training images, for activation maximization. Then the authors repeated this process to generate different images using activation maximization. They called it \emph{multifaceted feature visualization}. However, the generated images are noisy and not much diverse. Later, \citet{Nguyen_17a} used their plug and play model to generate comparatively diverse and more realistic images using Markov chain Monte Carlo based sampling approach.

Activation maximization is a good way to visualize the neuron of interest, but it does not consider the fact that real life images do not usually have high activations \citep{Nguyen_16a}. So, producing images which have very high activation value does not solve the problem entirely to understand what real life images are preferred by a neuron. 

So, to better understand the nature of a neuron in a deep neural network, we need an approach that can generate multiple natural looking images, and those images excite a given neuron in a certain activation range. To our best knowledge, this problem hasn't been addressed yet. Although authors in \citet{Nguyen_17a} generate more diverse samples for a single neuron compared to all the previous work, they do not address the issue mentioned above. There is no control over the activation value of the generated samples. Besides their Markov chain Monte Carlo based sampling approach is very slow. 

In the next section, we describe our approach to address the issue mentioned above.

\section{The inverse set of a neuron, and how to sample it}
\label{s:inverse-set}
\subsection{The inverse set of a neuron: definition}
We say an input \x\ is in the inverse set of a given neuron having a real-valued activation function $f$ if it satisfies the following two properties:
\begin{align}
  \label{e:feasiblity_equation}
  z_1\leq f(\x) \leq z_2 \quad \text{ and } \quad
  \x\text{ is a valid input}
\end{align}
where $z_1$, $z_2 \in \bbR$ are activation values of the neuron. \x\ being a valid input means the image features are in the valid range (say, pixel values in [0,1]) and it is a natural looking image.

For a simple model, the inverse set can be calculated analytically. For example, consider a linear model with logistic activation function $\sigma(\w^T\x+c)$ and all valid inputs to have pixel values between [0,1]. For $z_2=1$ (maximum activation value) and $0 < z_1 < z_2$, the inverse set will be the intersection of the half space $\w^T\x+c \geq \sigma^{-1}(z_1)$  and the [0,1] hypercube.

One way to look at this inverse set is as feasibility problem which has general form as:  
\begin{align}
  \centering 
  &\argmin_{\x} \hspace{1ex}  1 \hspace{3ex}\text{such that: set of constraints(\x)}  \quad
  \Leftrightarrow \quad \X = \{\x: \text{set of constraints(\x)} \}.
  \label{e:feasiblity_set}
\end{align}
In the case of deep neural networks, these constraints are nonlinear. So, we approximate the $\X$ with a sample $S$ that covers it in a representative way. The difficult part is to find an efficient algorithm to construct $S$. 

A simple way to do this is to select all the images in the training set that satisfy eq.~\eqref{e:feasiblity_equation}, but this may rule out all images. A neuron may ``like'' certain aspects of a training image without being sufficiently activated by it, or, in other words, the images that activate a given neuron need not look like any specific training image. Therefore, we need an efficient algorithm to sample the inverse set.

\subsection{Sampling the inverse set of a neuron: an optimization approach}
To generate the maximum activation image, the problem can be mathematically formulated as: 

\begin{equation}
  \centering 
  \argmax_{\x}  f(\x)+\calR(\x)
  \label{e:maxactivation }
\end{equation}
where real-valued function $f$ gives the activation value of the given neuron for an input image $\x$.  $\calR$ is a regularizer which makes sure that the generated image $\x$ looks like a real image. This is the same objective function used in \citet{Simony_14a}, \citet{Yosinski_15a}, \citet{Nguyen_16b}, and others, where,  $\calR$ is replaced by their hand-crafted regularizers. As mentioned in \citet{Nguyen_16b}, it mostly produces the same images for a given neuron. However, to construct the sample $S = \{\x_1,\cdots, \x_n  \}$ that covers the inverse set, generated images should be different from each other. So, we propose the following formulation to construct $S$ of size $n$ as a constraint optimization problem:

\begin{align}
  \centering
  \argmax_{\x_1,\x_2,\cdots, \x_n } \sum^n_{i,j=1}\norm{\x_i-\x_j}^2_2 \qquad
  \text{s.t.} \quad z_1 \le f(\x_1),\dots,f(\x_n)\le z_2.
  \label{e:samping_with_images}
\end{align}
The objective function makes sure that the samples are different from each other and also satisfy eq.~\eqref{e:feasiblity_equation}. However, this generates noisy-looking samples. To make them realistic we use an image generator network \G, which has been empirically shown to produce realistic images \citep{DosovitBrox16b} when a feature vector $\c$ is passed as an input. Then we get:
\begin{equation}
  \argmax_{\c_1,\c_2,\cdots, \c_n } \sum^n_{i,j=1}\norm{\G(\c_i)-\G(\c_j)}^2_2 \qquad
  \text{s.t.} \quad z_1 \le f(\G(\c_1)),\dots,f(\G(\c_n))\le z_2.
  \label{e:samplingforAlex}
\end{equation}
We observe that using Euclidean distances directly on the generated images $\G(\c)$ is very sensitive to small changes in their pixels. Instead, we compute distances on a low-dimensional encoding $\E(\G(\c))$ of the generated images, where \E\ is obtained from the first layers of a deep neural network trained for classification. Then we have our final formulation of the optimization problem over the $n$ samples $\G(\c_1),\dots,\G(\c_n)$:
\begin{equation}
\argmax_{\c_1,\c_2,\cdots, \c_n } \sum^n_{i,j=1}\norm{\E(\G(\c_i))-\E(\G(\c_j))}^2_2   \quad
\text{s.t.} \quad z_1 \le f(\G(\c_1)),\dots,f(\G(\c_n))\le z_2.
\label{e:samplingforAlex_code}
\end{equation}
Now to generate the samples, initialize $\c$ with random values and then optimize eq.~\eqref{e:samplingforAlex_code} using augmented Lagrangian~\citep{NocedalWright06a}. 

In theory eq.~\eqref{e:samplingforAlex_code} is enough to generate the sample $S$. But in practice, as the $n$ gets larger which is required to correctly sample the inverse set, eq.~\eqref{e:samplingforAlex_code} pose two issues: First, because of the quadratic complexity of the objective function over the number of samples $n$, it is computationally expensive to generate many samples. Second, since it involves optimizing all codes together, for larger $n$ it is not possible to fit all in the GPU memory. In the next section, we describe a much faster and less computationally expensive approach to create the inverse set.

\subsection{Sampling in feasible region}
\label{s:incr_sam}
\label{s:incr_sam}
In this section, we solve the problem for sampling the set $S$ described in eq.~\eqref{e:samplingforAlex_code} with a faster approach. For this, we apply two approximations.

First, we solve the problem in an inexact but good enough way. The sum-of-all-pairs objective is not a strict necessity; it is really a mechanism to ensure the diversity of the samples and coverage of the inverse set. We observe that this is already achieved by stopping the optimization algorithm once the samples enter the feasible set, by which time they already are sufficiently separated.

Second, we create the samples incrementally, $K$ samples at a time (with $K \ll n$). For the first $K$ samples (which we call \emph{seeds} ($\C_0$)) we optimize eq.~\eqref{e:samplingforAlex_code}, initializing the code vectors \c\ with random values and stopping as soon as all $K$ samples are in the feasible region. These samples are then fixed. The next $K$ samples are generated by the following equation:

\begin{align}
  \argmax_{\c_1,\c_2,\cdots, \c_K } &\sum^K_{i,j=1}\norm{\E(\G(\c_i))-\E(\G(\c_j))}^2_2  + 
  \sum^K_{i=1}\sum^{|C_0|}_{y=1}\norm{\E(\G(\c_i))-\E(\G(\c_y))}^2_2 &  \nonumber \\
  \text{s.t.}  &\qquad z_1 \le f(\G(\c_1)),\dots,f(\G(\c_K))\le z_2 \quad 
  \text{ and } \quad  \c_y\in \C_0.
  \label{e:feasible_region}
\end{align}
The first part of the equation ``$ \sum^K_{i,j=1}\norm{\E(\G(\c_i))-\E(\G(\c_j))}^2_2$'' is similar to that of eq.~\eqref{e:samplingforAlex_code}, means samples should be apart from each other. On the other hand, the second part of the equation ``$ \sum^K_{i=1}\sum^{|C_0|}_{y=1}\|\E(\G(\c_i))$ $-\E(\G(\c_y))\|^2_2 $'' makes sure that the generated samples should be far apart from the previous ones. The presence of constraints makes sure that generated samples stay in the feasible region.

Now to pick next $K$ samples, we initialize them to the previous $K$ samples ($\C_0$) and take a single gradient step in the augmented Lagrangian optimization of eq.~\eqref{e:feasible_region}. This gives $K$ new samples ($\G(\c_i)$) which we fix, and the process is repeated until we generate the desired $n$ samples. 

Note that, here we are not trying to optimize anything. We are using eq.~\eqref{e:feasible_region} to take steps inside the feasible region. It is like taking a random walk, but here we are taking steps inside the feasible region in a way that the next step gives the samples which are different from each other as well as from the one we already have. Another good part of this approach is that unlike the previous sampling approaches eq.~\eqref{e:feasible_region}, this approach allows us to pick samples in parallel which further increase the speed of sampling 
\footnote{ In most of our experiments $K =10 $ and $n=500$. It took almost $85$ gradient steps of eq.~\eqref{e:feasible_region} to generate the rest of  $490$ samples with $K=10$ seeds. These are just a few extra gradient steps than the minimum required $49$ gradient steps.} 

\subsection{Optimization details}
\label{s:opt_details}

As described above creating the inverse set of a neuron takes place in two steps:
\begin{itemize}
  \item Finding seeds in the feasible region. 
  \item Use these seeds to pick samples inside the feasible region. 
\end{itemize}
For finding seeds we optimize eq:~\eqref{e:samplingforAlex_code}, that can be transformed into:
\begin{align}
  \argmax_{\c_1,\c_2,\cdots, \c_K } \sum^K_{i,j=1}\norm{\E(\G(\c_i))-\E(\G(\c_j))}^2_2   \quad \nonumber \\
  \text{s.t.} \quad  z_2-f(\G(\c_i)) \leq \epsilon \quad  \forall i \in \{1,\cdots,K\} \text{ and} \quad  \epsilon= z_2-z_1.
  \label{e:samplingforAlex_code_eps}
\end{align}
This is a constraint optimization problem with inequality constraints. We can solve it by introducing slack variables ($s$), and then using bound-constrained augmented Lagrangian. So,  eq:~\eqref{e:samplingforAlex_code_eps} will transform to:
\begin{align}
\calL_{\text{seed}}(\C;\sss,\blambda,\mu) =\argmin_{\C } -&\sum^K_{i,j=1} \norm{\E(\G(\c_i))-\E(\G(\c_j))}^2_2\nonumber \\
 -&\sum^K_{i=1}\lambda_i (\epsilon_0 -z_2+ f(\G(\c_i))-s_i)\nonumber\\  
 + &\sum^K_{i=1} \frac{\mu}{2}(\epsilon_0 - z_2 + f(\G(\c_i))-s_i)^2  \nonumber \\
&\text{s.t.} \qquad s_i \geq 0 \quad \forall i \in \{1,\cdots,K\}.
\label{e:Aug_seeds}
\end{align}
Here, $\epsilon_0 = \frac{\epsilon}{2}$ , $\sss = \{s_1,s_2,\cdots,s_K\}$ and $\C= \{\c_1,\c_2,\cdots, \c_K\}$. The reason to make $\epsilon_0 = \frac{\epsilon}{2}$ is that once we have seeds($\C_0$), they should be in well inside the feasible region rather on the boundary. This helps generate samples to get started from the middle of the feasible region, which pushes them to both sides of the feasible region one towards the $z_2$ and other away from $z_2$.  It helps generated samples to cover the inverse set at much faster pace.

Now, to optimize eq:~\eqref{e:Aug_seeds}, initialize $\C$ by passing $K$ initial images $\X_0$ (we used random values) through the encoder($\E$) , $\blambda \in \bbR^K$ with zero column vector of size $K$ and $\mu \in \bbR$ with small scalar value $\mu_0 > 0$.

Now apply ``coordinate descent'' first in $\sss$, then in $\C$:
\begin{itemize}
  \item \textbf{In} $\sss$: 
    \begin{align*}
      &\argmin_{\sss} \calL_\text{seed}(\C;\sss,\blambda,\mu) \text{	s.t.} \quad s_i \geq 0 \quad \forall i \in K \\
      &\implies s^r_i = \text{max}(0,\epsilon_0 -z_2+ f(\G(\c_i))- \frac{1}{\mu_r}\lambda_i^r).
    \end{align*}
    Since $\calL_\text{seed}$ is a convex, separable quadratic form on $\sss$. 
  \item \textbf{In} $\calL_\text{seed}$: substitute the value of $\sss$ in $\calL_\text{seed}$ and then solve it approximately using gradient descent as unconstrained problem.
\end{itemize}
After each step of coordinate descent first update $\blambda$ then $\mu$ as follows:
\begin{itemize}
  \item $\lambda^{r+1}_i \leftarrow \text{max}(\lambda^{r}_i - \mu^r(\epsilon_0 -z_2+ f(\G(\c_i)),0)$.
  \item  $\mu^{r+1} \leftarrow \alpha\mu^r$ where  $\alpha\in \bbR$ and $\alpha> 1$.
\end{itemize}
We stop the optimization process when all $\c$ reach to the feasible region, means  
\begin{equation*}
  z_2- f(\G(\c_i))   \leq \epsilon \quad \forall i \in \{1,\cdots,K\}.
\end{equation*}
The codes $\C$ which we get after this optimization act as seeds($\C_0$) for doing sampling in the feasible region using eq:~\eqref{e:feasible_region}.

Let us say the value of augmented Lagrangian parameters at the end of optimization is $\blambda^*$ and $\mu^*$. 

In most of our experiments value of $\mu$ starts with $10$ and updated with a multiplication factor of $10$ that is  $\alpha = 10$. For each iteration of coordinate descent we run $100$ steps of gradient descent to approximately optimize $\calL_\text{seed}$ with $K =10$. It takes around $4$ iterations of coordinate descent to find the seeds in the feasible region which takes approximately $4$ minutes on NVIDIA Quadro P5000-16GB GPU, that we used for all our experiments.

Now, to pick samples in the feasible region using seeds we use eq:~\eqref{e:feasible_region}, and that can be transformed into:
\begin{align}
  \argmax_{\c_1,\c_2,\cdots, \c_K } \sum^K_{i,j=1}\norm{\E(\G(\c_i))-\E(\G(\c_j))}^2_2   +  \sum^K_{i=1}\sum^{|C_0|}_{y=1}\norm{\E(\G(\c_i))-\E(\G(\c_y))}^2_2   \nonumber \\
  \text{s.t.} \quad  z_2-f(\G(\c_i)) \leq \epsilon \quad  \forall i \in \{1,\cdots,K\} \text{ and} \quad  \epsilon= z_2-z_1.
  \label{e:feasible_region_eps}
\end{align}
As described above we need to take the gradient of eq.~\eqref{e:feasible_region_eps} with respect to $\C = \{\c_1,\c_2,\cdots, \c_K\}$ to generate rest of the samples. But eq.~\eqref{e:feasible_region_eps} contains inequality constraints so we transform it into bound-constrained augmented Lagrangian as we did for eq.~\eqref{e:samplingforAlex_code_eps}, but the augmented Lagrangian parameters $\blambda$ and $\mu$ are initialized with $\blambda^*$ and $\mu^*$ and never changed again throughout the sampling process. 

\begin{align}
  \calL_\text{sample}(\C;\C_0,\sss,\blambda^*,\mu^*) =\argmin_{\C }& -\sum^K_{i,j=1}\norm{\E(\G(\c_i))-\E(\G(\c_j))}^2_2   -  \sum^K_{i=1}\sum^{|C_0|}_{y=1}\norm{\E(\G(\c_i))-\E(\G(\c_y))}^2_2    \nonumber \\ 
  &-\sum^K_{i=1}  \lambda_i^* (\epsilon_0 -z_2+ f(\G(\c_i))-s_i) + \sum^K_{i=1} \frac{\mu^*}{2}(\epsilon_0 -z_2+ f(\G(\c_i))-s_i)^2 \nonumber\\
  &\text{s.t. }\quad s_i \geq 0 \quad \forall i \in \{1,\cdots,K\}.
  \label{e:Aug_feasibleregion}
\end{align}

Now to do the sampling we apply following steps:
\begin{enumerate}
  \item Initialize $\C$ with $\C_0$.
  \item \label{step2}Update $\sss$ as follows:
    \begin{equation*}
      s^r_i \leftarrow \text{max}(0,\epsilon_0 -z_2+ f(\G(\c_i))- \frac{1}{\mu^*}\lambda_i^*).
    \end{equation*}
  \item \label{step3}Take \textbf{single} gradient step of $\calL_\text{sample}$ with respect to $\C$ and update $\C$ to $\C^{\prime}$ as follows:
    \begin{equation*}
      \C^{\prime} \leftarrow \C - \beta \frac{\partial \calL_\text{sample} }{\partial \C} 
    \end{equation*}
    where $\beta \in \bbR$ is the step length.
  \item \label{step4} Replace $\C_0$ with  $\c^{\prime}_i$ which are in feasible region that is for which $z_2- f(\G(\c^{\prime}_i))   \leq \epsilon$. These $\G(\C_0)$ are the new samples. 
  \item \label{step5}Assign $\C'$ to $\C$.
  \item Repeat step~\ref{step2},~\ref{step3},~\ref{step4} and~\ref{step5} until all $n$ desired are not generated.
\end{enumerate}

Step~\ref{step3}, not only helps the algorithm to move inside the feasible region but it also provides a direction where the generated samples are different from the one we already have. The presence of constraint terms in eq.~\eqref{e:Aug_feasibleregion} play a very important role in the sampling process. In absence of the constraint the eq.~\eqref{e:Aug_feasibleregion} will be as below:
\begin{align*}
  \calL_\text{sample}(\C;\C_0) =\argmin_{\C }& -\sum^K_{i,j=1} \norm{\E(\G(\c_i))-\E(\G(\c_j))}^2_2- \sum^K_{i=1}\sum^{|C_0|}_{y=1} \norm{\E(\G(\c_i))-\E(\G(\c_y))}^2_2  
\end{align*}
which simply means the new codes $\c_i$ should be far apart from each other as well as from the one we already have ($\C_0$). This leads the gradient (step~\ref{step3}) to move the new samples ($\G(\C^{\prime})$) outside the feasible region. After few iterations, all the generated samples ($\G(\C^{\prime})$) will be out of the feasible region and there is no way to bring them back to the feasible region, which makes it impossible to generate a large number of samples. Infact , the constraint terms act as a penalty in case any of the $\G(\c_i^{\prime})$ goes outside the feasible region which pushes it back to the feasible region in next gradient update steps (step~\ref{step3}).

\subsection{Intersection of inverse-sets}
Our method also allows us to visualize the intersection of multiple inverse sets. This can be achieved by modifying the constraints in eq.~\eqref{e:samplingforAlex_code} as:

\begin{align}
  \argmax_{\c_1,\c_2,\cdots, \c_n }  \sum^n_{i,j=1}  &\norm{\E(\G(\c_i))  -  \E(\G(\c_j))}^2_2   \nonumber \\
  \text{s.t.} \quad z^{(1)}_1 \le f_1(\G(\c_1)),&  \dots,  f_1(\G(\c_n))\le z^{(1)}_2 , \nonumber \\
  \vdots   \nonumber  \\
  z^{(p)}_1 \le f_p(\G(\c_1)),&  \dots,  f_p(\G(\c_n))\le z^{(p)}_2
  \label{e:intrs_Alex_code}
\end{align}
and in ~\eqref{e:feasible_region} as follows:
\begin{align}
  \argmax_{\c_1,\c_2,\cdots, \c_K } \sum^K_{i,j=1} \norm{\E(\G(\c_i))-\E(\G(\c_j))}^2_2   &+
  \sum^K_{i=1}\sum^{|C_0|}_{y=1}\norm{\E(\G(\c_i))-\E(\G(\c_y))}^2_2   \nonumber \\
  \text{s.t.} \quad z^{(1)}_1 \le f_1(\G(\c_1)), \dots,&f_1(\G(\c_K))\le z^{(1)}_2, \nonumber \\
  \vdots  \nonumber  \\
  z^{(p)}_1 \le f_p(\G(\c_1)), \dots,&f_p(\G(\c_K))\le z^{(p)}_2  \text{and } \quad \c_y\in \C_0 
  \label{e:intrs_feasiblity_region}
\end{align}
where, $f_1, \cdots , f_p$ are real valued activation functions for different neurons and $(z^{(1)}_1,z^{(1)}_2),\cdots , (z^{(p)}_1,z^{(p)}_2)$ are corresponding activation values.

\begin{figure*}[ht!]
  \begin{tabular}{@{}c@{}c@{}}
    \rotatebox{90}{\hspace{.3ex}{act:[40,50]}}&
    \includegraphics*[width=.98\linewidth]{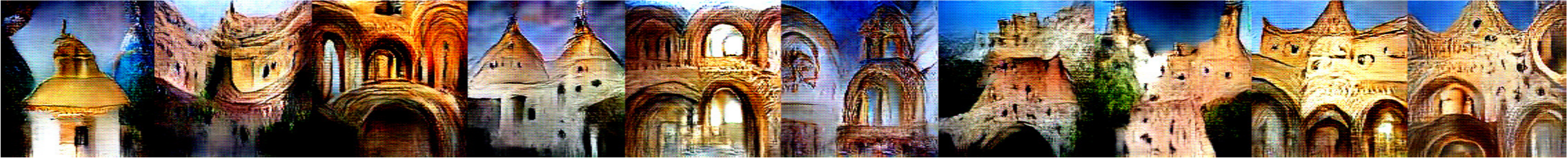} \\
    \rotatebox{90}{\hspace{.3ex}{act:[40,50]}}&
    \includegraphics*[width=.98\linewidth]{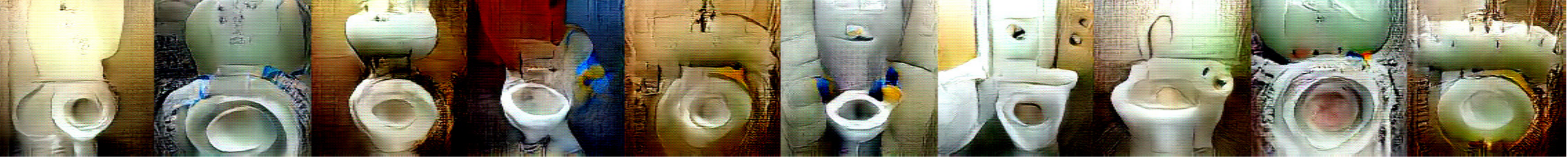}\\\\
    &{Inverse set Intersection}\\
    &\includegraphics*[width=.98\linewidth]{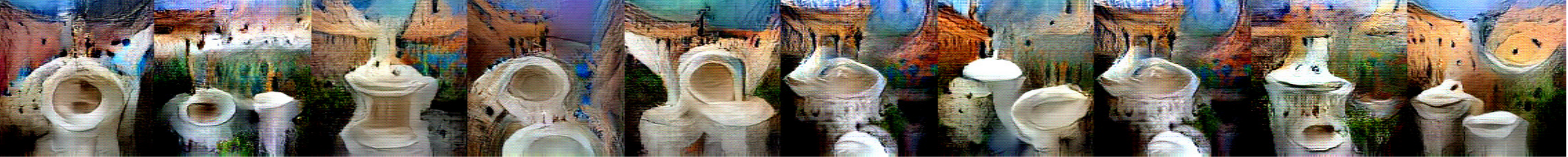}
  \end{tabular}
  \caption[caption]{First two rows show the samples generated by our sampling method for neuron number $664$ (represents \textbf{Monastery}) and $862$ (represents \textbf{Toilet seat}) respectively. Both neurons are from layer fc8 of CaffeNet \citep{Jia_14a}. Last row contains the samples from their intersection set. All three rows have activation range of [40,50]. } 
  \label{f:662_862_50_mix_short}

\end{figure*}

\begin{figure*}[ht!]
  \centering
  \begin{tabular}{@{}c@{}c@{}}
    \includegraphics*[width=1\linewidth]{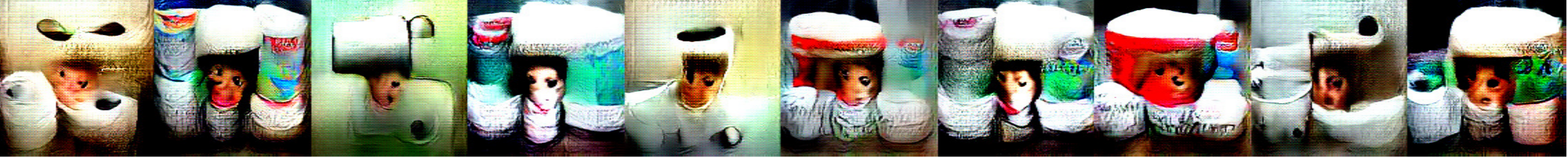} \\\\
    \includegraphics*[width=1\linewidth]{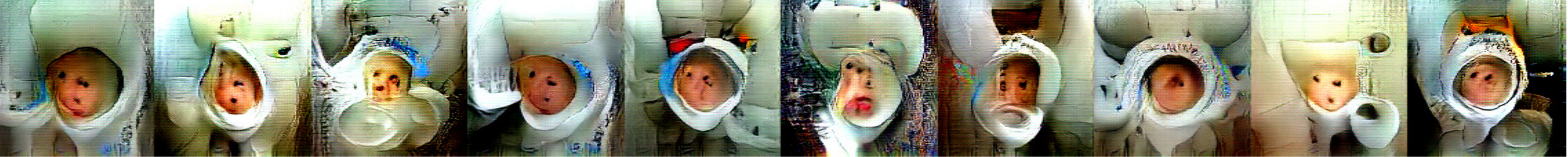}
  \end{tabular}
  \caption[caption]{ Above figure contains samples from the intersection of the inverse set of a hidden neuron and a neuron in the fc8 layer of CaffeNet \citep{Jia_14a}. The hidden neuron in both rows is neuron number $197$ in the conv5 layer, which is known to detect faces. The activation range for the hidden neuron is $[260,300]$. In the first row, the neuron from the fc8 layer is neuron number $1000$ (toilet paper class), and the activation range is $[20, 50]$. The generated samples have activation value in the range of $[260,300]$ for the hidden neuron and  $[20, 50]$ for the neuron number $1000$ in the fc8 layer.  In the second row the neuron in the fc8 layer is neuron number $862$ (toilet seat class), and activation range is $[40, 50]$. The generated samples have activation value in the range of $[260,300]$ for the hidden neuron and  $[40, 50]$ for the neuron number $862$ in the fc8 layer. } 
  \label{f:1000_862_197_mix_short}
\end{figure*}

Figure~\ref{f:662_862_50_mix_short} shows samples from the intersection of two different inverse sets, corresponding to neurons in the same layer. Equation~\eqref{e:intrs_feasiblity_region} can also be used to generate samples from inverse set intersection of neurons in different layers, as shown in fig.~\ref{f:1000_862_197_mix_short}. Generated samples have a face in the middle of the images. When these samples passed through the network, they excite both neurons in their corresponding activation range.

\section{Experiments}
\label{s:expts}

\begin{figure*}[t!]
  \centering
  \begin{tabular}{@{}c@{}c@{\hspace{1ex}}c@{}c@{}c@{}}
    \rotatebox{90}{\hspace{.1ex}{\#981}}&&
    \rotatebox{90}{\hspace{.1ex}{act:[50,60]}}&
    \rotatebox{90}{\hspace{3ex}{ours}}&
    \includegraphics*[width=.935\linewidth]{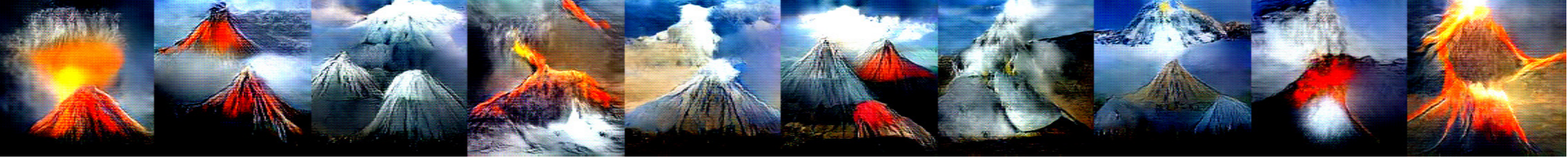}\\
    \rotatebox{90}{\hspace{3ex}{neuron}}&&&
    \rotatebox{90}{\hspace{3ex}{PPGN}}&
    \includegraphics*[width=.935\linewidth]{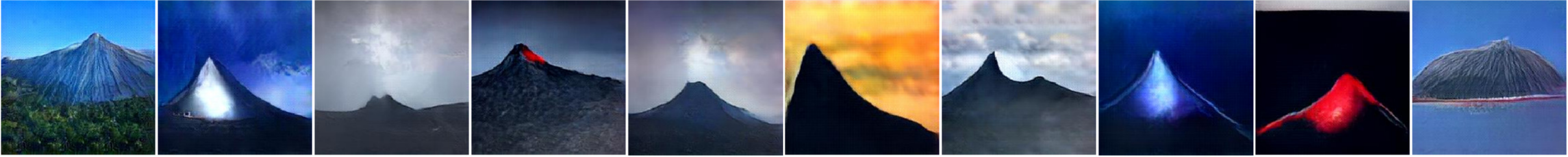}\\\\
    \rotatebox{90}{\hspace{.1ex}{\#947}}&&
    \rotatebox{90}{\hspace{.1ex}{act:[50,60]}}&
    \rotatebox{90}{\hspace{3ex}{ours}}&
    \includegraphics*[width=.935\linewidth]{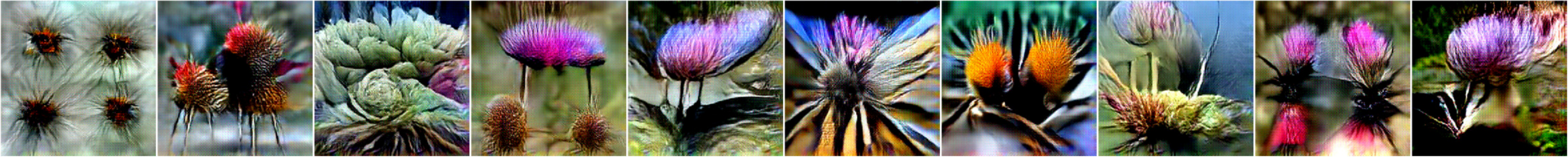}\\
    \rotatebox{90}{\hspace{3ex}{neuron}}&&&
    \rotatebox{90}{\hspace{3ex}{PPGN}}&
    \includegraphics*[width=.935\linewidth]{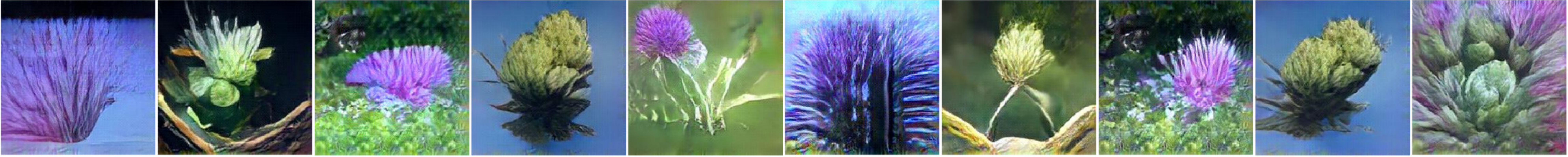}\\\\
    \rotatebox{90}{\hspace{,1ex}{\#1000}}&&
    \rotatebox{90}{\hspace{.1ex}{act:[40,50]}}&
    \rotatebox{90}{\hspace{3ex}{ours}}&
    \includegraphics*[width=.935\linewidth]{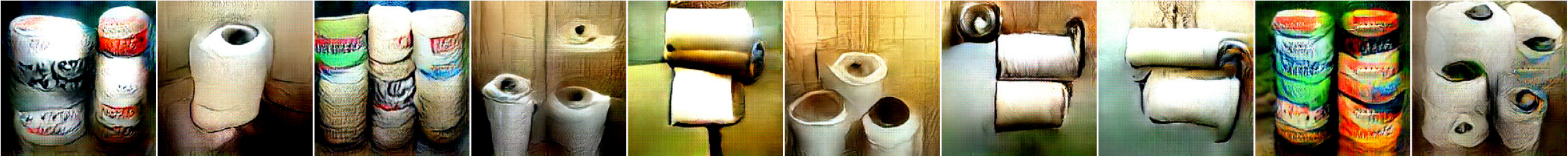}\\
    \rotatebox{90}{\hspace{3ex}{neuron}}&&&
    \rotatebox{90}{\hspace{3ex}{PPGN}}&
    \includegraphics*[width=.935\linewidth]{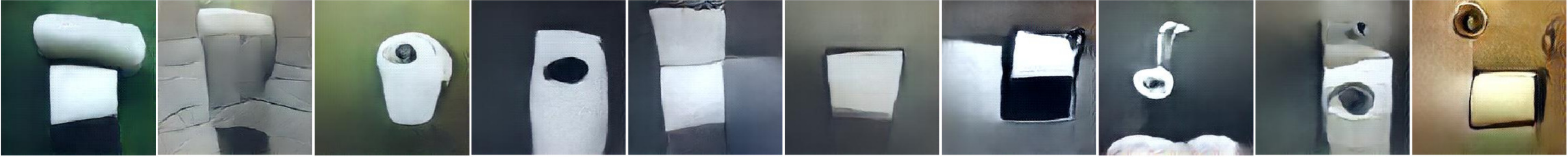}\\
  \end{tabular}
  \caption{The first, third and fifth row contains $10$ samples picked from $500$ samples generated by our sampling approach to cover the inverse set for the neuron number $981$ (represents volcano class), $947$ (represents cardoon class) and $1000$ (represents toilet paper class)  respectively. All three neurons are from layer fc8 of CaffeNet \citep{Jia_14a}. For the first row (volcano class) the activation range is [50,60], for the third (cardoon class) and the fifth row (toilet paper class) the range is [40,50]. 
  The second, fourth and sixth row shows samples generated for the same neurons by sampling approach from \citet{Nguyen_17a}. Activation range for the samples from \citet{Nguyen_17a} is not guaranteed to be in any fixed range like ours.}
  \label{f:fc8_ppgn_comparison}
\end{figure*}  

\begin{figure*}[ht!]
  \centering
  \begin{tabular}{@{}c@{}c@{}}
    \rotatebox{90}{\hspace{5.2ex}{\makebox[0ex][c]{act:[170,200]}}}&
    \includegraphics*[width=.98\linewidth]{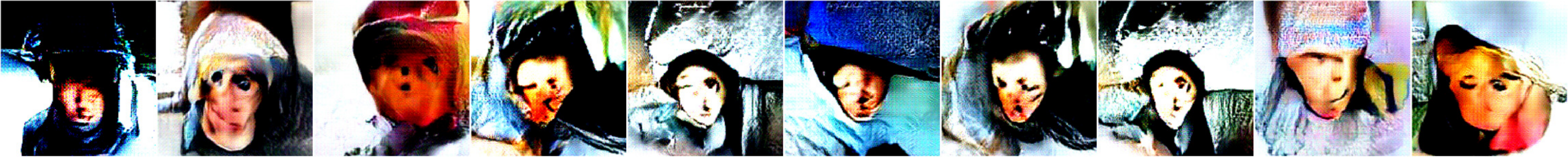}\\\\
    \rotatebox{90}{\hspace{5.2ex}{\makebox[0ex][c]{act:[260,300]}}}&
    \includegraphics*[width=.98\linewidth]{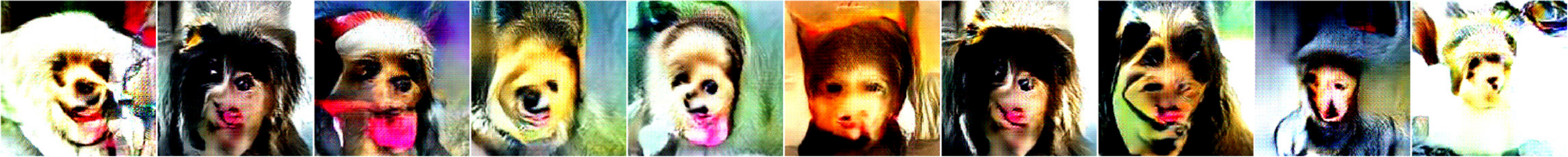}
  \end{tabular}
  \caption{Both rows contain $10$ samples out of $500$ samples which are generated to cover the inverse set for neuron number $197$  in the conv5 layer of CaffeNet \citep{Jia_14a}. This neuron is known to detect faces \citep{Yosinski_15a}. In the first row, the activation range is [$170,200$], and samples are mostly human faces. On the other hand, in the second row which has higher activation range( [$260,300$]), contains dog faces with fur. This shows the neuron prefers faces with fur compare to clean faces.} 
  \label{f:conv5_7_7_197_sample}
\end{figure*} 

\begin{table}[h!]
  \centering 
  \begin{tabular}{c|c@{\hspace{5ex}}c@{\hspace{5ex}}|@{\hspace{5ex}}c@{\hspace{5ex}}}
    \toprule
    {\textbf{\hspace{4ex}\textbf{Neuron in fc8}\hspace{4ex}}}&&Ours&PPGN~\citep{Nguyen_17a}\\
    \midrule
    981(volcano class) && 5.58 & 4.82\\
    947(cardoon class) && 6.00 & 4.99 \\
    1000(toilet paper class) && 5.83 & 4.79\\
    \bottomrule
  \end{tabular}
  \caption{Comparison between the  proposed method and PPGN; in terms of mean pairwise euclidean distance between the codes ($\E(\G(\c_i))$) of the generated samples (all 500 samples in fig.~\ref{f:fc8_ppgn_comparison}). Higher values mean generated samples are more diverse.}
  \label{t:numeric_comp}
\end{table}

\vspace{2ex}
\begin{figure*}[ht!]
  \centering
  \begin{tabular}{@{}c@{}c@{}}
    \rotatebox{90}{\hspace{.3ex}{act:[40,50]}}&
    \includegraphics*[width=.98\linewidth]{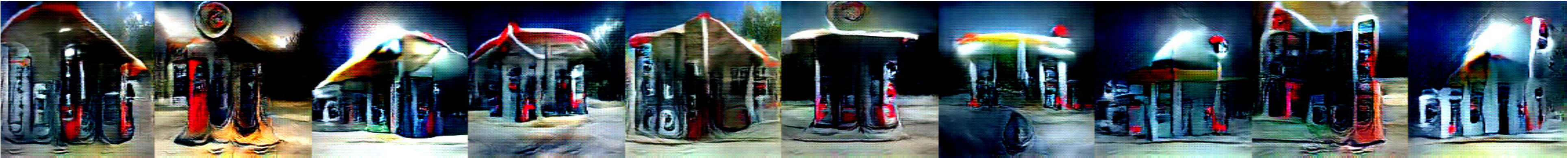}\\\\
    \rotatebox{90}{\hspace{.3ex}{act:[40,50]}}&
    \includegraphics*[width=.98\linewidth]{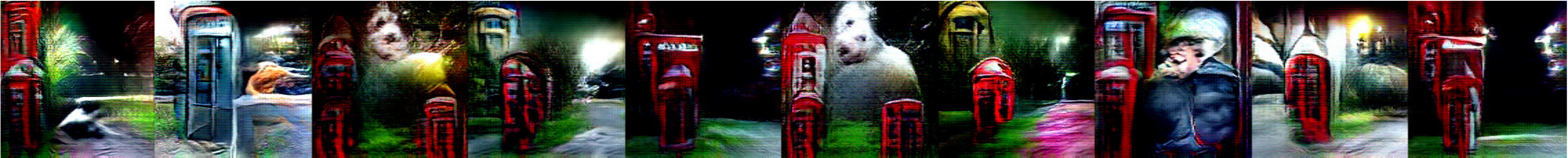}
  \end{tabular}
  \caption{Figure shows the results of our sampling process over a different network. The network is a variant of AlexNet trained on MIT Places dataset \citep{Zhou_14a}. The first row contains samples from the inverse set for neuron number $88$ (Gas station class ), and the second row contains samples from the inverse set for neuron number $141$ (Phone booth class). Both neurons are from the last layer before softmax, and the activation range is $[40,50]$ in both cases.}

  \label{f:fc8_mit_88_141_sample}
\end{figure*}

All previous works to understand the neural networks took CaffeNet \citep{Jia_14a} a minor variant of AlexNet \citep{Krizhev_12a} as the main subject. So, we also used a pre-trained CaffeNet \citep{Jia_14a} for our experiments, which had been trained on ImageNet dataset \citep{Deng_09a}. We use Matcovnet \citep{VedaldLenc15a} for all our experiments. Using previous papers  \citep{Nguyen_16a,Nguyen_17a}, naming convention we will also call last three fully connected layers in CaffeNet \citep{Jia_14a} as fc6, fc7, and fc8. fc8 is the last layer before softmax. In all our experiments \E\ is pre-trained CaffeNet \citep{Jia_14a}. However, the network has been shortened to the fc6 layer which is the first fully connected layer. The output of the \E\  is a vector of size $4096$. \G\ is a pre-trained generative network from \citet{Nguyen_17a} \footnote{\url{https://github.com/Evolving-AI-Lab/ppgn}}. \G\ \footnote{The only reason we use same network as~\citet{Nguyen_17a} is to do fair comparison of diversity in the generated images. Proposed method is independent of $\G$, so it can replaced with any state-of-the-art generative network.}
has been trained to generate images from a feature vector of size $4096$ more specifically output of fc6 layer of CaffeNet \citep{Jia_14a}. 

To do a fair comparison of our results with \citet{Nguyen_17a}, we pick the neuron $981$ which represents ``Volcano'' class. We run our algorithm for six times to generate $500$ samples to cover inverse set for this neuron corresponding to different activation levels. Figure~\ref{f:fc8_1_1_981_sample} shows results of this experiment. We pick $10$ samples from generated $500$ samples for each [$z_2,z_1$]. 

Unlike the previous visualization approaches \citep{Nguyen_17a,Nguyen_16a,Nguyen_16b} (second row of fig.~\ref{f:fc8_ppgn_comparison} ), the generated samples are far more diverse and rich in information. Our approach allows us to look at the samples which excite the neuron at different activation levels; this was not possible before. In doing so, it uncovers samples which have never been seen before. For instance, images which do not even contain volcano, instead contain lava flowing through water, but still excite the neuron. 

Figure~\ref{f:fc8_ppgn_comparison} shows some of the samples generated by our algorithm for a certain activation range with some other neurons and their comparison with \citet{Nguyen_17a}. To do a fair comparison with \citet{Nguyen_17a}, we generate samples as it is described in the supplementary section of the paper. We run 10 sampling chains conditioned on various classes each for 200 steps, to produce 2000 samples for each class. We picked $1$ sample from each $200$ steps to provide as much diversity as possible in the samples.  

Second row shows the generated samples for neuron number $947$ which represents ``Cardoon'' class for  $z_2=  50$ and $z_1=  40$. The generated samples not only have more diverse color distribution compared to the \citet{Nguyen_17a} (fourth row in the fig.~\ref{f:fc8_ppgn_comparison}) but also showing samples of different shapes and sizes, that can only be seen in real life images.  

To further test whether our sampling truly generates diverse samples or not, we pick a rather difficult class: ``Toilet paper'', neuron number $1000$. The reason for the difficulty is as the realistic looking samples cannot just differ by having a different color like in the case of ``Cardoon'' class; they all should have the white color. Sampling method should pick the samples which are of different shape or quantity. The fifth row in fig.~\ref{f:fc8_ppgn_comparison} shows our results. The generated samples are very diverse; they not only show just a toilet roll or toilet rolls hanged with a hanger but also show packed toilet rolls. In fact, samples contain packages with different shapes, labels, and even quantity. The first and ninth sample in the fifth row of fig.~\ref{f:fc8_ppgn_comparison} shows rolls packed in two packages but have altogether different packaging labels, while the third sample shows toilet rolls packed in 3 packages which also have different packaging label than other two. Our sampling method was even able to pick samples such that the number of toilet rolls is different, as shown in the second, fourth and tenth sample. These type of samples can only be seen in real life images thus showing how perfectly our sampling method covers the inverse set. On the other hand, samples generated by using PPGN (Plug and Play Generative Networks) \citep{Nguyen_17a} (sixth row of fig.~\ref{f:fc8_ppgn_comparison}) mostly contains rolls which are standalone or attached to a holder clearly not much diverse. This diversity also translates numerically, as shown in table~\ref{t:numeric_comp}.

We also tested our method for hidden neurons. We apply our algorithm for neuron number $197$ in the conv5 layer (last convolution layer in the CaffeNet \citep{Jia_14a}). This neuron is known to detect faces \citep{Yosinski_15a}. We first apply our approach with activation range $[200,170]$, most of the generated samples produce human faces as shown in the first row of fig.~\ref{f:conv5_7_7_197_sample}. But as we apply the same process with activation range $[300,260]$, dog's faces started to appear in the samples as shown in the second row of the fig.~\ref{f:conv5_7_7_197_sample}. Hence showing this neuron starts getting more activation towards the faces with more fur compared to that of clean faces. 

\textbf{Note:} We have created animations to show all the samples that are generated for different experiments in this paper.  Refer to author,s website for details about the animations.

\subsection{Need of encoder($\E$) in sampling process}

\begin{figure*}[t!]
  \centering
  \begin{tabular}{@{}c@{}}
    \includegraphics*[width=1\linewidth]{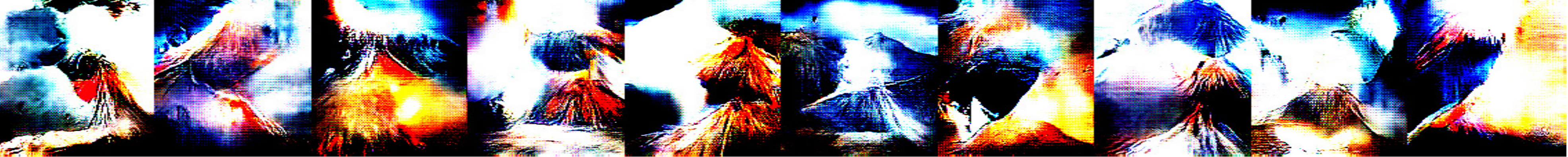}\\
    \includegraphics*[width=1\linewidth]{r_0_10_60_fc8_981.pdf}
  \end{tabular}
  \caption{All images are generated for the neuron number $981$ in the fc8 layer, which represents Volcano class. In both cases activation range is [50,60]. The first row samples are generated using eq.~\eqref{e:opt_with_codes} which looks really bad visually unlike the second row samples which are generated by using eq.~\eqref{e:samplingforAlex_code} that are much closer to real life image.} 
  \label{f:fc8_1_1_981_codes_fullopt}

\end{figure*}

Both in eq.~\eqref{e:samplingforAlex_code} and eq.~\eqref{e:feasible_region} samples are generated by taking $\ell_2$ distance in code space. Mathematically speaking, the objective function is just a distance between two vectors. But in our case, these vectors (codes $\c$) go through a nonlinear transform (first $\G$ and then $\E$) before $\ell_2$ distance is applied to them. Take instance of eq.~\eqref{e:samplingforAlex_code}:
\begin{align*}
  \centering
  \argmax_{\c_1,\c_2,\cdots, \c_K } \sum^K_{i,j=1}\norm{\E(\G(\c_i))-\E(\G(\c_j))}^2_2   \quad
  \text{s.t.} \quad z_1 \le f(\G(\c_1)),\dots,f(\G(\c_K))\le z_2.
\end{align*} 
However, one can argue why we need this transform as it is just a distance between two vectors. So, instead of involving encoder($\E$) and generator network($\G$) in eq.~\eqref{e:samplingforAlex_code}, we generated samples by optimizing directly over codes in the following way:
\begin{align}
  \centering
  \argmax_{\c_1,\c_2,\cdots, \c_K } \sum^K_{i,j=1}\norm{\c_i-\c_j}^2_2   \quad
  \text{s.t.} \quad z_1 \le f(\G(\c_1)),\dots,f(\G(\c_K))\le z_2.
  \label{e:opt_with_codes}
\end{align}  
Although, this seems to be a much cleaner approach, then eq.~\eqref{e:samplingforAlex_code} but this results in generating samples which are visually very bad as compare to eq.~\eqref{e:samplingforAlex_code} as shown in the first row of fig.~\ref{f:fc8_1_1_981_codes_fullopt}. Generated samples have large white spots, and they barely look close to the real-life images, unlike the samples generated by the eq.~\eqref{e:samplingforAlex_code} (second row in fig.~\ref{f:fc8_1_1_981_codes_fullopt}), which looks visually much better. 

The possible reason for this difference in the quality of the generated images is the presence of certain extreme values (large negative and positive values) in the code vector($\c$) when samples are generated using eq.~\eqref{e:opt_with_codes}. 
But when samples are generated using eq.~\eqref{e:samplingforAlex_code} the presence of the encoder($\E$) limits the values of the code vector $\c$ in a certain range, which results in more smooth and more natural-looking generated images. However, when codes have been optimized using eq.~\eqref{e:opt_with_codes} there is no restriction on what values of code vector ($\c$) can take, which make them to take certain extreme values (large negative and positive values). This leads to generating images that have big white spots and barely natural-looking images as shown in the first row of fig.~\ref{f:fc8_1_1_981_codes_fullopt}.

\subsection{Generating diverse samples from a single initial value }
To check the effect of term  $\sum^K_{i,j=1}\norm{\E(\G(\c_i))-\E(\G(\c_j))}^2_2  $ in eq.~\eqref{e:samplingforAlex_code} for generating diverse samples we perform following experiment. 

\begin{figure*}[t!]
  \centering
  \begin{tabular}{@{}l@{}l@{}}
    \rotatebox{90}{\hspace{.35ex}{act:[50,60]}}&
    \includegraphics*[width=.98\linewidth]{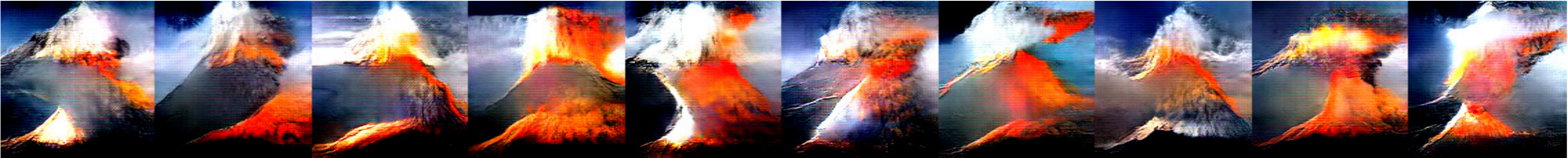}\\
    \rotatebox{90}{\hspace{.3ex}{act:[50,60]}}&
    \includegraphics*[width=.98\linewidth]{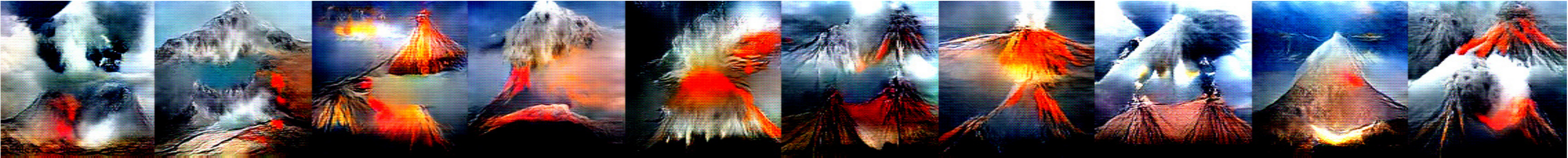}
    
  \end{tabular}
  \caption{ All images are generated for the neuron number $981$ in the fc8 layer, which represents Volcano class. In both cases activation range is [50,60]. The first row samples are generated using eq.~\eqref{e:no_obj}  such that all $\c$ are initialized with the same random vector. Generated samples look very similar unlike the second row samples that are comparatively more diverse and are generated by using eq.~\eqref{e:samplingforAlex_code} with all $\c$ are initialized with the same value as in the first row.}
  \label{f:single_init_60_20_1_1_981_rad_10_explore}
\end{figure*}

First we modify the  eq.~\eqref{e:samplingforAlex_code} as follows:
\begin{equation}
  \centering
  \argmax_{\c_1,\c_2,\cdots, \c_K  }  \quad 1     \quad
  \text{s.t.} \quad z_1 \le f(\G(\c_1)),\dots,f(\G(\c_K))\le z_2.
  \label{e:no_obj}
\end{equation}
Then we initialize all $\c$ from same random vector and optimize eq.~\eqref{e:no_obj} using coordinate as described in section~\ref{s:opt_details}. First row of fig.~\ref{f:single_init_60_20_1_1_981_rad_10_explore} shows the generated samples. All samples are very little diverse. Then we repeat the same experiment to generate samples using eq.~\eqref{e:samplingforAlex_code} initializing all $\c$ with the same value as above. The results are shown in  second row of fig.~\ref{f:single_init_60_20_1_1_981_rad_10_explore}. As expected generated are comparatively diverse, showing the need of the term  $\sum^K_{i,j=1}\norm{\E(\G(\c_i))-\E(\G(\c_j))}^2_2   $ in eq.~\eqref{e:samplingforAlex_code} for generating diverse samples. 

\subsection{Proposed fast sampling process is a good approximation of Eq~\ref{e:samplingforAlex_code} }
Equation~\eqref{e:samplingforAlex_code} is actually enough to generate samples which are different from each other and look like real life images. But as described above it is very slow and may not be be possible due to memory constraints for large number of samples. So we try to approximate this whole process by our modified sampling process as described in section~\ref{s:opt_details}. There is one more approximation we apply while optimizing eq.~\eqref{e:samplingforAlex_code}. We stop our optimization as soon as all our samples reach to the feasible region. 

To test how good is our approximation we perform following experiment:

\begin{enumerate}
  \item First we generate $n$ samples instead of $K$ ($K \ll n$) samples at a time, using eq.~\eqref{e:samplingforAlex_code}  as follows:
    \begin{align}
      \argmax_{\c_1,\c_2,\cdots, \c_n } \sum^n_{i,j=1}\norm{\E(\G(\c_i))-\E(\G(\c_j))}^2_2   \quad
      \text{s.t.} \quad z_1 \le f(\G(\c_1)),\dots,f(\G(\c_n))\le z_2
      \label{e:all_in_onego}
    \end{align}  

  \item Next, using same initial values (picked first $K$ initial images from $n$ used in above experiment), we first generate $K$ seeds using eq.~\eqref{e:samplingforAlex_code}. But instead of stopping the optimization as soon as we reach to feasible region we optimize eq.~\eqref{e:samplingforAlex_code} all the way. After that we generate rest of the $n-K$ samples using eq.~\eqref{e:feasible_region} as described in section~\ref{s:opt_details}.

  \item Last we generate all $n$ samples with the same parameters and same initial values for $K$ seeds in above step using sampling process described in section~\ref{s:opt_details}.  
\end{enumerate}

\begin{figure*}[ht!]
  \centering
  \begin{tabular}{@{}c@{}c@{}}
    \rotatebox{90}{\hspace{7.8ex}{act:[50,60]}}&
    \includegraphics*[width=.98\columnwidth]{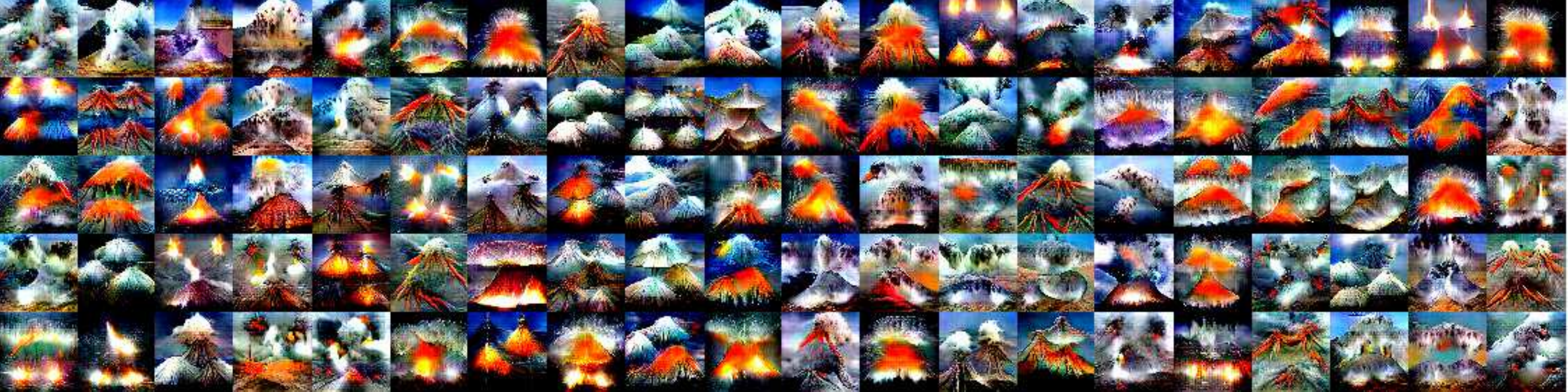} \\
    \\
    \rotatebox{90}{\hspace{7.8ex}{act:[50,60]}}&
    \includegraphics*[width=.98\columnwidth]{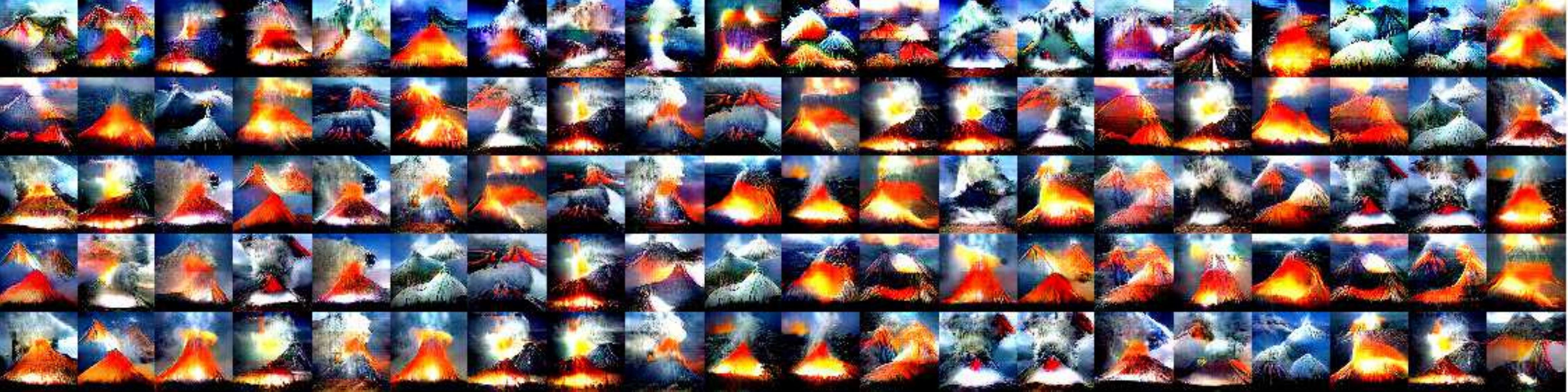}\\
    \\
    \rotatebox{90}{\hspace{7.8ex}{act:[50,60]}}&
    \includegraphics*[width=.98\columnwidth]{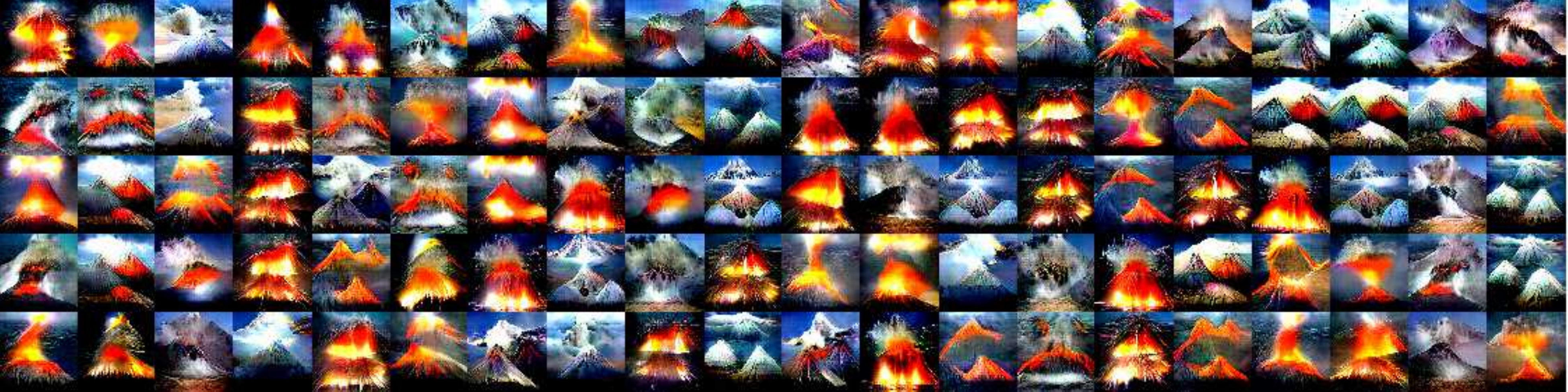}\\

  \end{tabular}
  \caption[caption]{In all three cases neuron in concern is number 981 in layer fc8 of  CaffeNet \citep{Jia_14a}  with activation range $[50,60]$. \emph{First block: } All $100$ samples (first 5 rows) are generated by fully optimizing eq.~\eqref{e:all_in_onego}. \emph{Second block: } the first row ( sixth row in total) represents the seeds which are generated by fully optimizing using eq.~\eqref{e:samplingforAlex_code}. That is, unlike proposed sampling approach we didn't stop optimization as soon as all samples were in feasible region. Rest of the samples are generated by our fast sampling method (eq.~\eqref{e:feasible_region}). \emph{Third block: } the first row ($11^{th}$ row in total) represents the seeds which are generated as per the optimization details described in section~\ref{s:opt_details} and rest of the images are generated by our fast sampling method (eq.~\eqref{e:feasible_region}).  From the figure, it is pretty clear that the proposed sampling approach approximate eq.~\eqref{e:all_in_onego} pretty well in much fewer iterations. } 
  \label{f:981_60_full}

\end{figure*}

For these three experiments, the total number of samples generated are $n = 100$ and number of seeds are  $K= 10$. The total number of gradient steps took to optimize eq.~\eqref{e:samplingforAlex_code} for experiment in 1 was around $13000$. Compare to this in experiment  2, the total number of gradient steps required for eq.~\eqref{e:samplingforAlex_code} were around $800$ and rest $n-K$ samples were generated under of $35$ steps. Whereas the last experiment took only around $400$ gradient steps which take almost 5-6 minutes over a GPU and rest $n-K$ samples were also generated under $35$ steps. Figure~\ref{f:981_60_full} shows results of this experiment. After looking at the results it is pretty clear that the proposed sampling process does a good job at approximating the solution of eq.~\eqref{e:all_in_onego} in much fewer iterations.     

\section{Discussion}
Admittedly, the goal of understanding what a neuron in a deep neural network may be representing is not a well defined problem. It may well be that a neuron does represent a specific concept, but one which is very difficult to grasp for a human; or that one should look at what a group of neurons may be representing. That said, for some neurons their preferred response does correlate well with intuitive concepts or classes, such as the volcano or monastery examples we give. Our approach is to characterize a neuron's preference by a diverse set of examples it likes, which is something that people sometimes do in order to explain a subjective concept to each other. Also, it may be possible to extract specific concepts from this set of examples using data analysis techniques.

\section{Conclusion}
In conclusion, we propose a very simple, yet effective generalized approach to understand the neurons in a deep neural network: an approach that does not involve activation maximization or feature inversion, but instead characterizes the region of input space around different activation values of the neuron of interest. Thus overcoming the shortcomings of current visualization approaches and providing a great deal of understanding what parts of an image are responsible for impacting the activation of a neuron. This eventually helps us to have a much better understanding about the nature of a neuron in deep neural networks. 

We also provide a simpler, faster, and hyper-parameter free sampling method which generates far more diverse samples than previously proposed methods. Our sampling method is also very generalized;  just by modifying the constraints, it can also be used for high dimensional sampling in other domains. Although we performed our experiments with only image classification models, our approach for creating inverse sets can also be applied to deep neural networks in other domains, as well as to other complex machine learning models.

%

\end{document}